\documentclass[letterpaper]{article}  % DO NOT CHANGE THIS

\usepackage{ifthen}  % ifthenelse macro

\newcommand\type{preprint}

\newcommand{\ifsubmission}[2]{\ifthenelse{\equal{\type}{submission}}{#1}{#2}}
\newcommand{\iffinal}[2]{\ifthenelse{\equal{\type}{final}}{#1}{#2}}
\newcommand{\ifpreprint}[2]{\ifthenelse{\equal{\type}{preprint}}{#1}{#2}}

\ifsubmission{}{\iffinal{}{\ifpreprint{}{\PackageError{}{Unknown type}{}}}}
\ifsubmission{%
  \usepackage[submission]{aaai23}     % DO NOT CHANGE THIS
}{%
  \usepackage[]{aaai23}               % DO NOT CHANGE THIS
}
\usepackage{times}                    % DO NOT CHANGE THIS
\usepackage{helvet}                   % DO NOT CHANGE THIS
\usepackage{courier}                  % DO NOT CHANGE THIS
\usepackage[hyphens]{url}             % DO NOT CHANGE THIS
\usepackage{graphicx}                 % DO NOT CHANGE THIS
\usepackage{natbib}                   % DO NOT CHANGE THIS AND DO NOT ADD ANY OPTIONS TO IT
\usepackage{caption}                  % DO NOT CHANGE THIS AND DO NOT ADD ANY OPTIONS TO IT
\usepackage{amsmath}   % math facilities (e.g., \lVert)
\usepackage{array}     % table alignment
\usepackage{bm}        % bold symbols (e.g., \boldsymbol)
\usepackage{booktabs}  % professional-quality tables
\usepackage{graphbox}  % graphics alignment
\usepackage{letterspace}  % reduce space between characters
\usepackage{lineno}    % line numbers for submission
\usepackage{ltxcmds}   % LaTeX utility macros (e.g., \ltx@ifpackageloaded)
\usepackage{tikz}
\usepackage{xcolor}    % color functionality (e.g., \textcolor)

\ifpreprint{\usepackage[backref=page]{hyperref}}{}

\ifsubmission{\linenumbers}{}

\nocopyright

\newcommand\highlight[1]{\textcolor{red}{\textbf{#1}}}
\newcommand\todo[1]{\highlight{\ifthenelse{\equal{#1}{}}{TODO}{TODO: #1}}}
\makeatletter\newcommand{\IfPackageLoaded}[3]{\ltx@ifpackageloaded{#1}{#2}{#3}}\makeatother
\newcommand\mailto[1]{\IfPackageLoaded{hyperref}{\href{mailto:#1}{#1}}{#1}}
\newcommand\narrowtt[1]{\textls[-30]{\texttt{#1}}}

\makeatletter
\@ifundefined{acro}{%
  \DeclareRobustCommand\SMC{%
    \ifx\@currsize\normalsize\small\else
    \ifx\@currsize\singlesize\small\else
    \ifx\@currsize\small\@setfontsize\small\@viiipt{9}\else  % customized
    \ifx\@currsize\footnotesize\@setfontsize\small\@viiipt{9}\else  % customized
    \ifx\@currsize\large\normalsize\else
    \ifx\@currsize\Large\large\else
    \ifx\@currsize\LARGE\Large\else
    \ifx\@currsize\scriptsize\tiny\else
    \ifx\@currsize\tiny\tiny\else
    \ifx\@currsize\huge\LARGE\else
    \ifx\@currsize\Huge\huge\else
    \small\SMC@unknown@warning
    \fi\fi\fi\fi\fi\fi\fi\fi\fi\fi\fi
  }
  \newcommand\SMC@unknown@warning{\@latex@warning{%
      \string\SMC: unrecognized text font size
      command---using \string\small}}
  \newcommand\textSMC[1]{{\SMC #1}}
  \newcommand\acro[1]{\textSMC{#1}\@}}{}
\makeatother

\newcommand\CHW{\acro{CHW}}
\newcommand\CIFARTEN{\acro{CIFAR-10}}
\newcommand\CW{\acro{CW}}
\newcommand\BIM{\acro{BIM}}
\newcommand\FGSM{\acro{FGSM}}
\newcommand\ReLU{\acro{ReLU}}
\newcommand\RGB{\acro{RGB}}

\DeclareMathOperator{\sign}{sign}
\DeclareMathOperator*{\argmin}{argmin}

\title{Deviations in Representations \\ Induced by Adversarial Attacks}
\author{
  Daniel Steinberg,\!\textsuperscript{\rm 1}
  Paul Munro\textsuperscript{\rm 2}
}
\affiliations{
  \textsuperscript{\rm 1} Intelligent Systems Program, University of Pittsburgh \\
  \textsuperscript{\rm 2} School of Computing and Information, University of Pittsburgh \\
  {\mailto{das178@pitt.edu}}, {\mailto{pwm@pitt.edu}}
}

\IfPackageLoaded{hyperref}{
  \ifsubmission{}{
    \hypersetup{
      pdfinfo={
        Title={Deviations in Representations Induced by Adversarial Attacks},
        Author={Daniel Steinberg, Paul Munro},
        TemplateVersion={2023.1}
      }
    }
  }
}

\begin{document}

\maketitle

\begin{abstract}

Deep learning has been a popular topic and has achieved success in many areas. It has drawn the
attention of researchers and machine learning practitioners alike, with developed models deployed to
a variety of settings. Along with its achievements, research has shown that deep learning models are
vulnerable to adversarial attacks. This finding brought about a new direction in research, whereby
algorithms were developed to attack and defend vulnerable networks. Our interest is in understanding
how these attacks effect change on the intermediate representations of deep learning models. We
present a method for measuring and analyzing the deviations in representations induced by
adversarial attacks, progressively across a selected set of layers. Experiments are conducted using
an assortment of attack algorithms---on the \CIFARTEN{} dataset---with plots created to visualize
the impact of adversarial attacks across different layers in a network.

\scalebox{0.96}[0.96]{Code is available at
  \mbox{\ifsubmission{\url{https://anonymized/for/submission}}%
    {\url{https://github.com/dstein64/adv-deviations}}}}.

\end{abstract}

\section{Introduction}

It has been shown that carefully crafted adversarial instances can deceive deep learning
models~\cite{szegedy_intriguing_2014}. This prompted further research to (1)~develop new attack
techniques, (2)~defend against instances created with such methods, and (3)~explore and characterize
the properties of adversarial attacks. Here we're interested in the latter task, seeking to
understand how adversarial instances effect change on the hidden layers of their target networks.

\paragraph{Our Contribution} We present a method for measuring and analyzing the deviations in
representations induced by adversarial instances. Experiments are conducted with an assortment of
attack algorithms, using two separate distance metrics for measuring deviations in representations.
Our plots show the transitional effect on representations induced by adversarial attacks.

\section{Preliminaries}

There are various ways to generate adversarial instances. A typical approach starts with an input
$x$ and synthesizes a small additive perturbation $\Delta x$ for which $x + \Delta x$ would deceive
a neural network---whereas a human would not perceive much difference between $x$ and $x + \Delta
x$.

Here we briefly cover the three adversarial attack methods that we incorporate into our experiments.
In addition to the $x$ and $\Delta x$ notation mentioned earlier, $y$ represents a ground truth
class label and $J$ is a loss function for a neural network---which has already been trained in this
scenario.

\textbf{Fast Gradient Sign Method~(\FGSM{})}~\cite{goodfellow_explaining_2015} generates an
adversarial perturbation $\Delta x$ = $\epsilon \cdot \sign(\nabla_x J(x, y))$, which is in the
approximate direction of the loss function gradient. The $\sign$ function transforms each element to
$-1$, $0$, or $1$, with the result then scaled by $\epsilon$. Thus, the $L_\infty$ norm of $\Delta
x$ is bounded by $\epsilon$.

\textbf{Basic Iterative Method~(\BIM{})}~\cite{kurakin_adversarial_2017} applies \FGSM{}
iteratively. At each step, $x^{adv}_{t} = x^{adv}_{t-1} + \alpha \cdot \sign(\nabla_x
J(x^{adv}_{t-1}, y))$, with initialization $x^{adv}_0 = x$. On each iteration, the $L_\infty$ norm
of the latter addition term is bounded by $\alpha$, and clipping at each step can be used to
constrain the final $x^{adv}$ to an $\epsilon$-neighborhood of $x$.

\textbf{Carlini \& Wagner (\CW{})}~\cite{carlini_towards_2017} constructs an adversarial
perturbation using gradient descent to solve $\Delta x = \argmin_{\delta} (\lVert\delta\rVert_p + c
\cdot f(x + \delta))$. Function $f$ is one for which $f(x + \delta) \leq 0$ if and only if the
attack is successful on the target classifier; their $f_6$ formulation was found most effective.
Positive constant $c$ can be determined using binary search. A box constraint on $x + \delta$ is
satisfied by clipping or change of variables. $p$ specifies which norm is used.

\section{Method}
\label{sec:method}

\begin{figure*}[tb]
  \centering
  \hspace{-0.13in}
  \begin{tikzpicture}
    \draw (0, 0) node[inner sep=0]%
      {\includegraphics[width=0.95\textwidth]{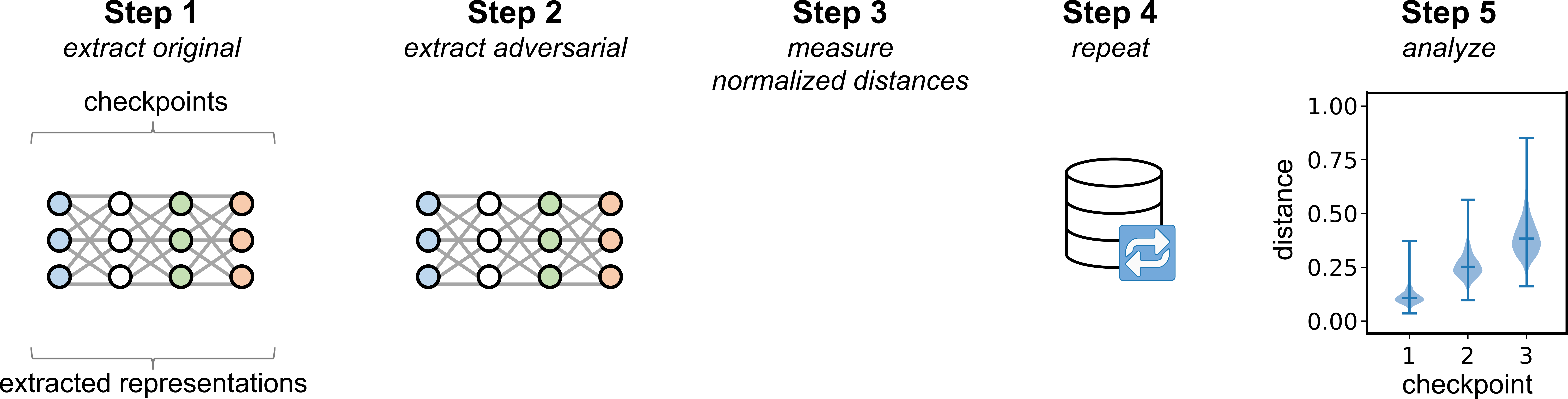}};
    \draw (0.65, 0.3) node {\large \textsf{1}: $d(\phi_1, \phi_1')/N_1$};
    \draw (0.65, -0.4) node {\large \textsf{2}: $d(\phi_2, \phi_2')/N_2$};
    \draw (0.65, -1.1) node {\large \textsf{3}: $d(\phi_3, \phi_3')/N_3$};
    \draw (-8.3, -0.45) node {\LARGE $x$};
    \draw (-7.79, 0.34) node {\large \textsf{1}};
    \draw (-6.48, 0.34) node {\large \textsf{2}};
    \draw (-5.83, 0.34) node {\large \textsf{3}};
    \draw (-7.75, -1.3) node {\large $\phi_1$};
    \draw (-6.45, -1.3) node {\large $\phi_2$};
    \draw (-5.77, -1.3) node {\large $\phi_3$};
    \draw (-4.38, -0.38) node {\LARGE $x'$};
    \draw (-3.83, 0.34) node {\large \textsf{1}};
    \draw (-2.51, 0.34) node {\large \textsf{2}};
    \draw (-1.86, 0.34) node {\large \textsf{3}};
    \draw (-3.81, -1.31) node {\large $\phi_1'$};
    \draw (-2.51, -1.31) node {\large $\phi_2'$};
    \draw (-1.83, -1.31) node {\large $\phi_3'$};
  \end{tikzpicture}
  \caption{
    Method illustration.
    \textbf{Step~1}:~Input $x$ is passed through a neural network, with representation vectors
    $\phi_1$, $\phi_2$, and $\phi_3$ extracted from selected checkpoints \textsf{1}, \textsf{2}, and
    \textsf{3}.
    \textbf{Step~2}:~Input $x'$\!---an adversarially perturbed version of $x$---is passed through
    the network, with representation vectors $\phi_1'$, $\phi_2'$, and $\phi_3'$ extracted from the
    same checkpoint locations used earlier.
    \textbf{Step~3}:~For each checkpoint, distance function $d$ (e.g., Euclidean) is used to measure
    the distance between corresponding representations extracted for $x$ and $x'$\!. The values are
    scaled using checkpoint-specific normalization constants, discussed further in
    Section~\ref{sec:method}.
    \textbf{Step~4}:~The earlier steps, one through three, are repeated for a dataset of inputs.
    \textbf{Step~5}:~The results are analyzed, showing the distribution of deviations induced by an
    adversarial attack at the selected checkpoints of the neural network.
  }
  \label{fig:method}
\end{figure*}
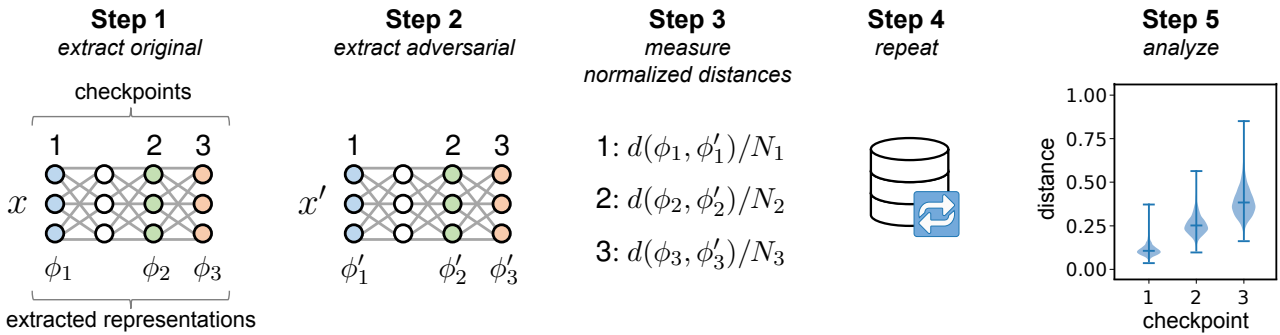

To inspect the deviations in representations induced by adversarial attacks, we start with a dataset
of original (non-attacked) instances, and an attacked version of the same dataset. For the neural
network that's targeted---or some other network, e.g., if the context is transferability---we select
a set of checkpoints, locations in the network for which we'll extract representation vectors.

By passing an original instance $x$ through the network, we extract representation vector $\phi_i$
at checkpoint $i$. We repeat this for the adversarial counterpart $x'$\!, extracting $\phi'_i$.
Next, we measure the distance $d(\phi_i, \phi_i')$ between the vectors. While we don't specify a
particular distance function $d$, our experiments use Euclidean distance and cosine distance.

The volume occupied by representations can differ at separate layers of a neural network. Therefore,
it's important to normalize our distance measurements prior to comparing them across checkpoints. We
follow the approach of \citeauthor[Section~4]{mahendran_understanding_2015}. Our earlier formulation
$d(\phi_i, \phi_i')$ becomes $d(\phi_i, \phi_i') / N_i$. Normalization constant $N_i$ (for
checkpoint $i$) is the average pairwise distance between representation vectors---computed using
$d$---across a sample of instances.

We calculate the normalized distances across all network checkpoints for the full dataset of
original and attacked instances (limited in our experiments to include only pairings where the
attack was successful). This provides a sample of deviations at each checkpoint, for which the
distributions can be analyzed. We use violin plots, but other techniques could also provide insight.
The method is illustrated in Figure~\ref{fig:method}.

\section{Experiments}
\label{sec:experiments}

\subsection{Experimental Settings}

Our experiments utilized the \CIFARTEN{} dataset~\cite{krizhevsky_learning_2009}, comprised of
60,000 $\text{32}{\times}\text{32}$ \RGB{} images---with a designated split into 50,000 training
images and 10,000 test images---across 10 classes. Using the training data, we fit a neural network
classifier that follows the \citeauthor{kuangliu_kuangliupytorch-cifar_2021} ResNet-18
architecture\footnote{This differs in filter counts and depth from the ResNet-20 architecture that
was used for \CIFARTEN{} in the original ResNet paper~\cite{he_deep_2016}.} with 11,173,962
parameters. With pixel values scaled by $\text{1}/\text{255}$ to be between zero and one, the
network was trained for 100 epochs using Adam~\cite{kingma_adam:_2015} for optimization, with the
training data augmented via (1)~random horizontal flipping and (2)~random crop sampling on images
padded with four pixels per edge. The resulting model had accuracy of 92.67\% on the test data.

\begin{figure*}[tb]
  \centering
  \includegraphics[width=0.98\textwidth]{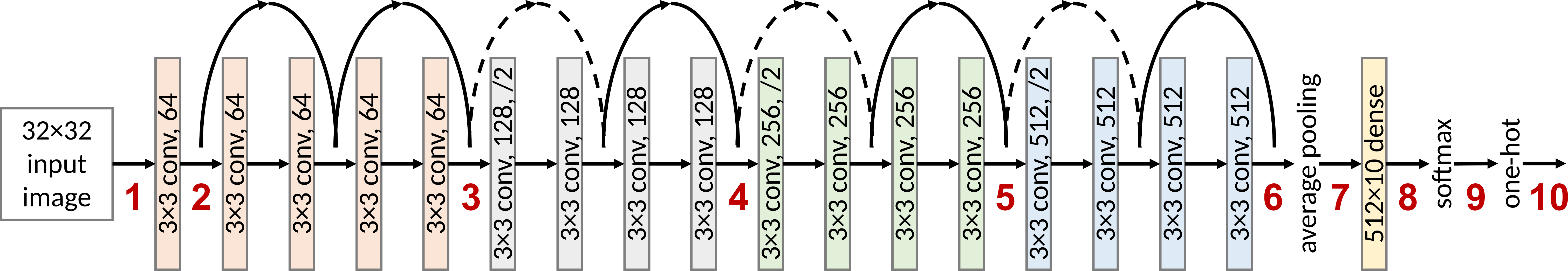}
  \caption{
    Model architecture and checkpoints. Red labels indicate where numbered checkpoints were placed
    in the ResNet-18 network. The first checkpoint captures the network's input prior to its
    subsequent processing. Solid arrow skip connections perform identity mapping, and the dashed
    arrow skip connections use pointwise convolution and stride to transform inputs to the necessary
    shapes for addition. The text in the colored rectangles indicates (1)~filter shapes, counts, and
    stride when present (e.g., ``/2" for stride of two), for convolutions, and (2) the weight matrix
    shape for the dense layer. Convolutional layers all include rectified linear unit (\ReLU{})
    activation functions. Where applicable, checkpoints are placed after the application of skip
    connection addition operations.
  }
  \label{fig:architecture}
\end{figure*}

We use 10 checkpoints throughout the model for extracting representations.
Figure~\ref{fig:architecture} displays the model architecture, highlighting the locations of the
numbered checkpoints. The shape and dimensionality of representations at each checkpoint are shown
in Table~\ref{table:shapes}.

\begin{table}[h]
  \caption{
    Shape and dimensionality of representations at the checkpoints used for our experiments. Shapes
    are reported channels-first, followed by height then width (\CHW{} format).
  }
  \label{table:shapes}
  \centering
  \begin{tabular}[b]{ccc}
    \toprule
    Checkpoint(s) & Shape & Dimensionality \\
    \midrule
    $1$ & $3 \times 32 \times 32$ & $3{,}072$ \\
    $2, 3$ & $64 \times 32 \times 32$ & $65{,}536$ \\
    $4$ & $128 \times 16 \times 16$ & $32{,}768$ \\
    $5$ & $256 \times 8 \times 8$ & $16{,}384$ \\
    $6$ & $512 \times 4 \times 4$ & $8{,}192$ \\
    $7$ & $512$ & $512$ \\
    $8, 9, 10$ & $10$ & $10$ \\
    \bottomrule
  \end{tabular}
\end{table}

\subsubsection{Adversarial Attacks}

We utilized the \texttt{cleverhans} library~\cite{papernot_technical_2018} to generate untargeted
adversarial attacks for the 9,267 correctly classified test images. Attacks were conducted with the
\FGSM{}, \BIM{}, and \CW{} algorithms. The attacked images were clipped between zero and one for all
attacks, and quantized to 256 discrete values---whereby each image could be represented in 24-bit
\RGB{} space---for \FGSM{} and \BIM{}. We did not quantize the \CW{}-attacked images, as doing so
would essentially revert the attack\footnote{The original \CW{} paper~\cite{carlini_towards_2017}
addresses the issue with a greedy search process that restores the attack quality one pixel at a
time---an approach that we did not utilize.}\!.

For \FGSM{}, $\epsilon$ was set to $\text{3}/\text{255}$ for a maximum perturbation of three
intensity values out of 255 for each pixel on the unnormalized data. Model accuracy was 16.80\% on
the 9,267 attacked images from the test dataset. That is, the attack succeeded at a rate of 83.20\%.

Our attacks generated with \BIM{} used 10 iterations with $\alpha = \text{1}/\text{255}$ and
$\epsilon = \text{3}/\text{255}$. This limits the per-step maximum perturbation to one unnormalized
intensity value per pixel, ultimately clipped to a maximum perturbation of three intensity values.
The resulting model accuracy on the attacked images was 0.29\% (i.e., an attack success rate of
99.71\%).

We used \CW{} with an $L_2$ norm distance metric along with the package's default parameters---five
binary search steps, a learning rate of 0.005, and up to 1,000 iterations. Accuracy after attack was
0.00\%, a perfect attack success rate.

\begin{figure}[ht]
  \begin{center}
    {
      \hspace{-25pt}
      \renewcommand{\arraystretch}{2.04}
      \newcommand\imgwidth{0.095\columnwidth}
      \newcommand\colwidth{0.44cm}
      \newcommand\shift{\hspace{-7pt}}
      \begin{tabular}{%
          r%
          p{\colwidth}%
          p{\colwidth}%
          p{\colwidth}%
          p{\colwidth}%
          p{\colwidth}%
          p{\colwidth}%
          p{\colwidth}%
        }
        Original\shift{} &
        \includegraphics[align=c,width=\imgwidth]{%
          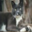} &
        \includegraphics[align=c,width=\imgwidth]{%
          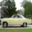} &
        \includegraphics[align=c,width=\imgwidth]{%
          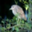} &
        \includegraphics[align=c,width=\imgwidth]{%
          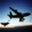} &
        \includegraphics[align=c,width=\imgwidth]{%
          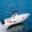} &
        \includegraphics[align=c,width=\imgwidth]{%
          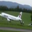} &
        \includegraphics[align=c,width=\imgwidth]{%
          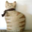} \\
        \FGSM{}\shift{} &
        \includegraphics[align=c,width=\imgwidth]{%
          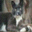} &
        \includegraphics[align=c,width=\imgwidth]{%
          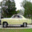} &
        \includegraphics[align=c,width=\imgwidth]{%
          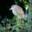} &
        \includegraphics[align=c,width=\imgwidth]{%
          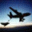} &
        \includegraphics[align=c,width=\imgwidth]{%
          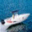} &
        \includegraphics[align=c,width=\imgwidth]{%
          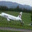} &
        \includegraphics[align=c,width=\imgwidth]{%
          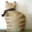} \\
        \BIM{}\shift{} &
        \includegraphics[align=c,width=\imgwidth]{%
          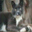} &
        \includegraphics[align=c,width=\imgwidth]{%
          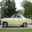} &
        \includegraphics[align=c,width=\imgwidth]{%
          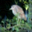} &
        \includegraphics[align=c,width=\imgwidth]{%
          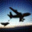} &
        \includegraphics[align=c,width=\imgwidth]{%
          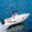} &
        \includegraphics[align=c,width=\imgwidth]{%
          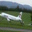} &
        \includegraphics[align=c,width=\imgwidth]{%
          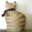} \\
        \CW{}\shift{} &
        \includegraphics[align=c,width=\imgwidth]{%
          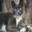} &
        \includegraphics[align=c,width=\imgwidth]{%
          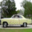} &
        \includegraphics[align=c,width=\imgwidth]{%
          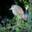} &
        \includegraphics[align=c,width=\imgwidth]{%
          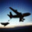} &
        \includegraphics[align=c,width=\imgwidth]{%
          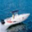} &
        \includegraphics[align=c,width=\imgwidth]{%
          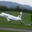} &
        \includegraphics[align=c,width=\imgwidth]{%
          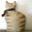}
      \end{tabular}
    }
  \end{center}
  \caption{
    Example \CIFARTEN{} images before and after adversarial perturbation. The top row has the
    original images, followed by three rows corresponding to \FGSM{}, \BIM{}, and \CW{} attacks.
    Images were randomly selected from the 9,267 test images that were correctly classified without
    perturbation (i.e., the set of images for which attacks were generated). The original image
    classes from left to right are \narrowtt{cat}, \narrowtt{automobile}, \narrowtt{bird},
    \narrowtt{airplane}, \narrowtt{ship}, \narrowtt{airplane}, and \narrowtt{cat}.
  }
  \label{fig:attacked_images}
\end{figure}

Figure~\ref{fig:attacked_images} shows examples of randomly selected \CIFARTEN{} images and their
adversarially perturbed counterparts.

\subsubsection{Measuring Distances}

As a next step we extract representations at the 10 selected checkpoints. This is conducted for all
9,267 test images that were correctly classified originally. We then repeat the extraction for the
adversarially perturbed counterparts of these same images. With representations extracted for both
original and adversarially attacked images, for each image we measure distances between the original
representations and corresponding adversarial counterparts. For our experiments we use two metrics
for measuring the distance between vectors $u$ and $v$, Euclidean distance $\lVert u - v \rVert$ and
cosine distance $1 - (u \cdot v)/(\lVert u \rVert \lVert v \rVert)$.

As we mentioned earlier, it's important to normalize the measured distances. We calculate
normalization constants using the 9,267 test images that were correctly classified initially. For
each checkpoint we extract the corresponding representations and then compute the average pairwise
distance across all ${\text{9}{,}\text{267} \choose \text{2}} = \text{42}{,}\text{934}{,}\text{011}$
pairs, to serve as divisors for scaling the originally calculated distances. This is done separately
for Euclidean and cosine metrics. That is, the originally measured Euclidean distances are
normalized by scaling constants calculated using Euclidean distance, and the cosine distances are
scaled with normalization constants determined using cosine distance.

\subsection{Results}

We just discussed the details of how we measured the distances between representations of original
images and their adversarial counterparts. To analyze the results, violin plots were employed to
visualize the distribution of distance measurements so that we could see how adversarial attacks
induce deviations progressively throughout the network.

\begin{figure*}[t]
  \centering
  \includegraphics[width=0.98\linewidth]{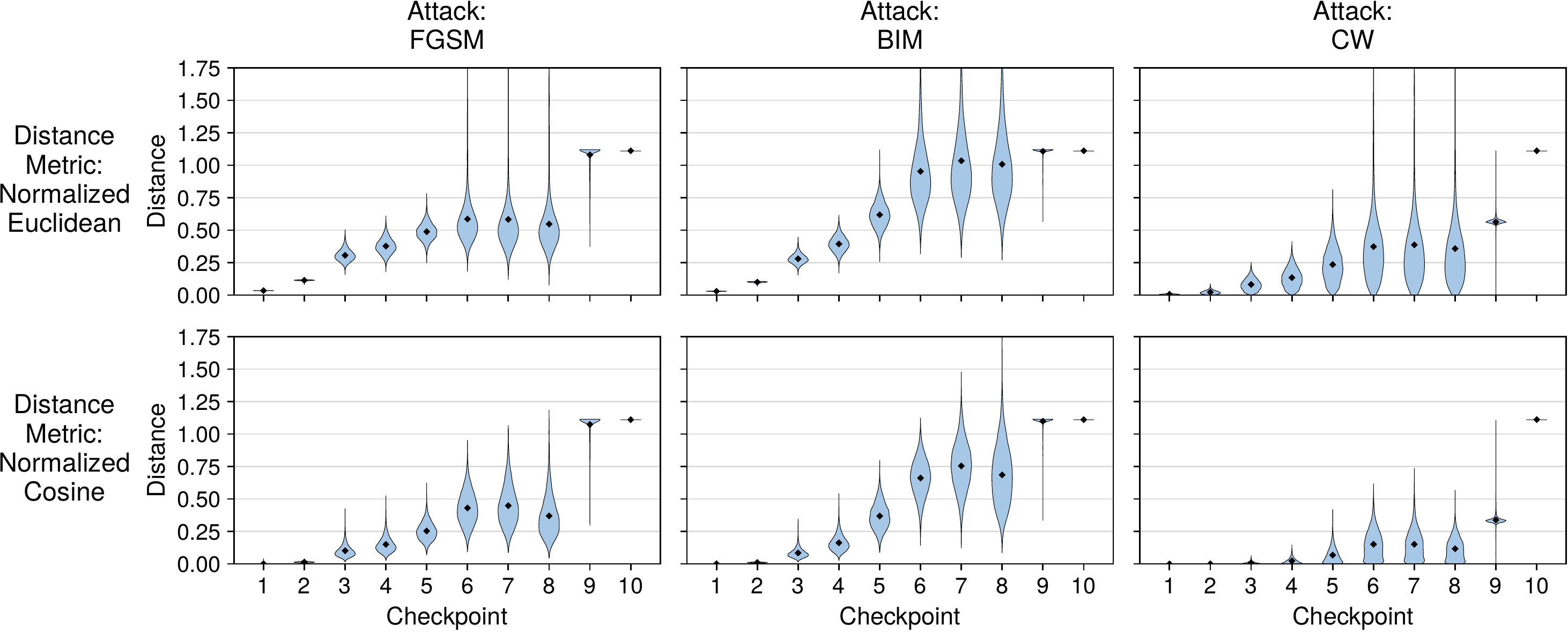}
  \caption{
    Violin plots showing the distribution of deviations in representations across selected
    checkpoint layers, induced by adversarial attacks. The figure subplots each correspond to a
    specific distance metric (indicated by the leftmost labels) and a specific attack algorithm
    (indicated by the header labels). Distance measurements were normalized. The diamond-shaped
    markers show sample means. We generated the plots using
    \texttt{matplotlib}~\cite{hunter_matplotlib_2007}, with the default settings of 100 points for
    Gaussian kernel density estimation and Scott's Rule for calculating estimator bandwidth.
  }
  \label{fig:plot}
\end{figure*}

Figure~\ref{fig:plot} shows the plots, which were generated across the three adversarial attacks and
two distance metrics described earlier. For each attack algorithm, our analysis is limited to the
subset of images that could be successfully attacked from the 9,267 correctly classified test
images---7,710 images for \FGSM{}, 9,240 for \BIM{}, and 9,267 for \CW{}. Because we only consider
succesfully attacked images, the deviations at checkpoint 10 are constant---the distance between two
non-equal one-hot vectors is always the same.

\section{Discussion}

\paragraph{Trend} Something that stands out is that the average deviation in
representations---induced by adversarially perturbing inputs---primarily increases as a function of
depth (i.e., normalized distances tend to be larger for higher numbered checkpoints). This matches
the general expectation we had prior to running the experiments. Interestingly, the pattern does not
hold for checkpoint 8, where the average distance was lower than it was at checkpoint 7. Checkpoint
8 occurs after applying the linear portion of a densely connected layer, whereas the earlier
checkpoints are chiefly positioned after the application of a convolution and \ReLU{} activation
function (the exceptions are checkpoint 1, which captures the input prior to subsequent network
processing, and checkpoint 7, which follows a pooling operation).

\paragraph{Attacks} Relative to the other attacks, it appears that \BIM{} induces the smoothest
transition in deviations from input to output. For \FGSM{} there is a noticeable jump in normalized
distances from checkpoint 8 to 9. A similar condition is present for \CW{}, but occurs between
checkpoint 9 and 10. For the configurations of attacks we considered, \CW{} led to the smallest
change in intermediate representations for original versus adversarially perturbed images.

\paragraph{Distance Metrics} The first row's plots, which use Euclidean distance, are similar to the
plots on the second row, which use cosine distance. The average normalized distances are always
higher for the Euclidean metric than for cosine, not counting checkpoint 10 where the means are the
same. This might be related to the property that two non-equal vectors in the same direction have
zero cosine distance between them, but positive Euclidean distance. For vectors of slightly
different direction, cosine distance would be small, whereas Euclidean distance could be large.

\section{Conclusion}

We set out to inspect how adversarial perturbations impact neural network representations. Our
proposed technique considers the normalized representation distances at selected checkpoints between
original images and their adversarially perturbed counterparts. Experiments conducted on \CIFARTEN{}
data using an assortment of attack algorithms allowed us to visualize the deviations in
representations induced by adversarial attacks. We observed different behavior---in the way
normalized distances transition from input to output throughout the network---across the
configurations of attacks we considered.

% **************************************
% * References
% **************************************

{
  % The AAAI formatting guidelines permit font size for references to be reduced to 9 point with 10
  % point linespacing if the paper would otherwise exceed the number of allowed pages.
  \fontsize{9}{10}\selectfont
  \bibliography{paper}
}

% Add References link to Contents. This has to be after \biliography, as the section is specified
% with that command.
\addcontentsline{toc}{section}{References}

\end{document}